\pdfoutput=1

\documentclass[11pt]{article}

\usepackage[final]{acl}

\usepackage{times}
\usepackage{latexsym}

\usepackage[T1]{fontenc}

\usepackage[utf8]{inputenc}

\usepackage{microtype}

\usepackage{inconsolata}

\usepackage{multirow}
\usepackage{graphicx}
\usepackage{enumitem}
\usepackage{cleveref}
\usepackage{tcolorbox}

\title{\textit{``We Demand Justice!''}:\\ Towards Social Context Grounding of Political Texts}

\author{Rajkumar Pujari \and Chengfei Wu \and Dan Goldwasser\\
  Purdue University, West Lafayette, USA\\
  \texttt{\{rpujari,wu1491,dgoldwas\}@purdue.edu}\\}

\begin{document}
\maketitle
\begin{abstract}
Political discourse on social media often contains similar language with opposing intended meanings. For example, the phrase \textit{thoughts and prayers}, is used to express sympathy for mass shooting victims, as well as satirically criticize the lack of legislative action on gun control. Understanding such discourse fully by reading only the text is difficult. However, knowledge of the social context information makes it easier. 

We characterize the social context required to fully understand such ambiguous discourse, by grounding the text in real-world entities, actions, and attitudes. We propose two datasets that require an understanding of social context and benchmark them using large pre-trained language models and several novel structured models. We show that structured models, explicitly modeling social context, outperform larger models on both tasks, but still lag significantly behind human performance. Finally, we perform an extensive analysis, to obtain further insights into the language understanding challenges posed by our social grounding tasks.
\end{abstract}

\section{Introduction}\label{sec:introduction}
\begin{figure}
    \centering
    \includegraphics[width=190pt]{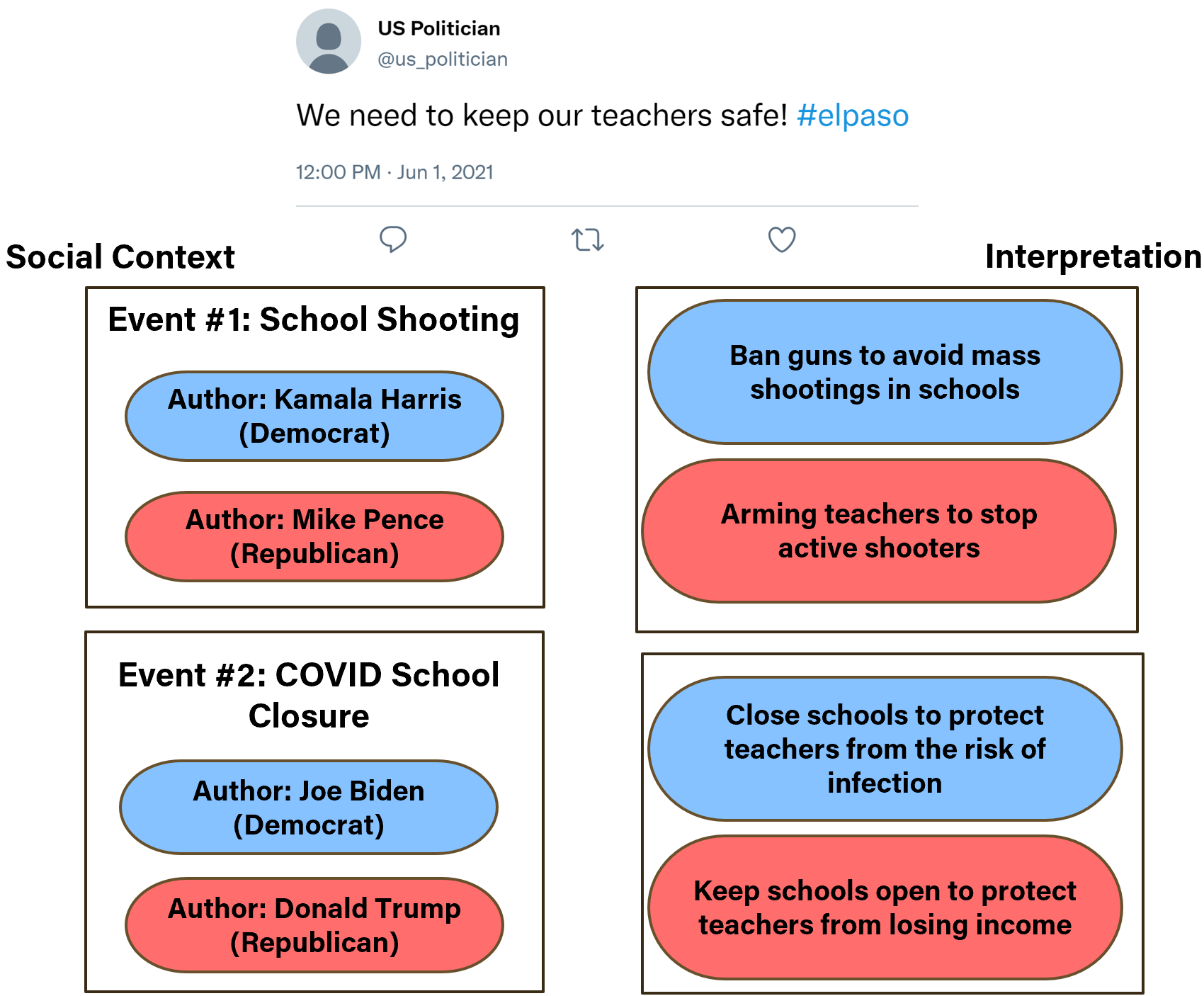}
    \caption{An example of varied \textit{intended meanings} behind the same political message depending on the Author and Event in context}
    \label{fig:intro_example}
\end{figure}

Over the past decade, micro-blogging websites have become the primary medium for US politicians to interact with general citizens and influence their stances for gaining support. As a result, politicians from the same party often coordinate the phrasing of their social messaging, to amplify their impact \citep{human_message_politics2011,Weber2021AmplifyingIT}. Hence, repetitive, succinct phrases, such as \textit{``Thoughts and Prayers''}, are extensively used, although they signal more nuanced stances. Moreover, the interaction among politicians from opposing parties often leads to messaging phrased similarly, but signaling opposing real-world actions. For example, \textit{`Thoughts and Prayers'}, when used by Republicans, expresses condolences in mass shooting events, but when used by Democrats conveys an \textit{angry} or \textit{sarcastic} tone as a call for action demanding \textit{``tighter gun control measures''}. Similarly, \cref{fig:intro_example} shows contrasting interpretations of the phrase ``\textit{We need to keep our teachers safe}!'' depending on different speakers and in the context of different events. 

\begin{table*}[htb]
\centering
\resizebox{400pt}{!}{
\begin{tabular}{llll|ll}
\hline
\multicolumn{4}{c|}{\textbf{Tweet Target Entity and Sentiment}}                                                                                                                                                                         & \multicolumn{2}{c}{\textbf{Vague Text Disambiguation}}                                                                                                                                                                                                                                                                                                                                                                                                                                                                                                                                                                                             \\ \hline
\multicolumn{4}{l|}{\begin{tabular}[c]{@{}l@{}}\textbf{Tweet:} As if we needed more evidence. \#kavanaugh\end{tabular}} & \multicolumn{2}{l}{\textbf{Vague Text:} First, but not the last.}                                                                                                                                                                                                                                                                                                                                                                                                                                                                                                                                                                                           \\
\multicolumn{4}{l|}{\textbf{Event:} Kavanaugh Supreme Court Nomination}                                                                                                                                                                          & \multicolumn{2}{l}{\begin{tabular}[c]{@{}l@{}}\textbf{Event:} US withdraws from Paris climate agreement that \\ enforces environmental targets after three years\end{tabular}}                                                                                                                                                                                                                                                                                                                                                                                                                                                                              \\
\multicolumn{4}{l|}{\textbf{Author:} Earl Blumenauer (Democrat Politician)}                                                                                                                                                                       & \multicolumn{2}{l}{\textbf{Author Party:} Republican}                                                                                                                                                                                                                                                                                                                                                                                                                                                                                                                                                                                                       \\
\multicolumn{4}{l|}{\begin{tabular}[c]{@{}l@{}}\textbf{Targets:} Brett Kavanaugh (negative), Julie Swetnick (positive) \\ Christine Ford (positive), Deborah Ramirez (positive)\end{tabular}}                                                                       & \multicolumn{2}{l}{\begin{tabular}[c]{@{}l@{}}\textbf{Disambiguation:} The withdrawal from the Paris climate agreement\\ is the first step of many to come for the Trump administration. It will\\ not be the last, as more positive changes are sure to follow.\end{tabular}}                                                                                                                                                                                                                                                                                                                                                                            \\
\multicolumn{4}{l|}{}                                                                                                                                                                                                                   & \multicolumn{2}{l}{\multirow{9}{*}{\begin{tabular}[c]{@{}l@{}}\textbf{Incorrect Disambiguations:} \\1) Joe Biden's inauguration marks the first day of a new era of progress\\  and prosperity, lasting positive changes are coming. (Incorrect Event)\\ 2) The Paris Climate Agreement withdrawal is the first of many\\ backward steps this Trump administration is sure to take in destroying\\ our environment. (Incorrect Stance)\\ 3) This is the time for America to move forward and make progress\\ without being held back by a global agreement that doesn't serve\\ our interests. (Doesn't match the vague text)\end{tabular}}} \\ \cline{1-4}
\multicolumn{2}{c|}{\textbf{Target Task Data Statistics}}                                                                      & \multicolumn{2}{c|}{\textbf{Vague Text Data Statistics}}                                                        & \multicolumn{2}{l}{}                                                                                                                                                                                                                                                                                                                                                                                                                                                                                                                                                                                                                               \\ \cline{1-4}
Unique Tweets                                               & \multicolumn{1}{l|}{$865$}                                         & \multicolumn{1}{l}{Unique Vague Texts}                                        & $93$                                         & \multicolumn{2}{l}{}                                                                                                                                                                                                                                                                                                                                                                                                                                                                                                                                                                                                                               \\
Positive Targets                                            & \multicolumn{1}{l|}{$1513$}                                        & Positive Examples                                         & $739$                                        & \multicolumn{2}{l}{}                                                                                                                                                                                                                                                                                                                                                                                                                                                                                                                                                                                                                               \\
Negative Targets                                            & \multicolumn{1}{l|}{$1085$}                                        & Negative Examples                                         & $2217$                                        & \multicolumn{2}{l}{}                                                                                                                                                                                                                                                                                                                                                                                                                                                                                                                                                                                                                               \\
Neutral Targets                                              & \multicolumn{1}{l|}{$784$}                                         & Total Examples                                            & $2956$                                        & \multicolumn{2}{l}{}                                                                                                                                                                                                                                                                                                                                                                                                                                                                                                                                                                                                                               \\
Non-Targets                                                 & \multicolumn{1}{l|}{$2509$}                                        & Number of Events                                          & $9$                                          & \multicolumn{2}{l}{}                                                                                                                                                                                                                                                                                                                                                                                                                                                                                                                                                                                                                               \\
Total Data Examples                                         & \multicolumn{1}{l|}{$5891$}                                        &  Hard Test Examples                                           & $180$                                           & \multicolumn{2}{l}{}                                                                                                                                                                                                                                                                                                                                                                                                                                                                                                                                                                                                                               \\
Number of Events                                            & \multicolumn{1}{l|}{$3$}                                           &                                                           &                                            & \multicolumn{2}{l}{}                                                                                                                                                                                                                                                                                                                                                                                                                                                                                                                                                                                                                               \\ \hline
\end{tabular}
}
\caption{Examples of Annotated Datasets and their statistics}
\label{tab:data_stats}
\end{table*}

Humans familiar with the stances of a politician and, possessing knowledge about the event from the news, can easily understand the \textit{intended meaning} of political phrases. However, automatically understanding such language is challenging. Our main question in this paper is - \textit{\textbf{Can an NLP model find the right meaning?}} From a linguistic perspective, we follow the distinction \citep{bach2008pragmaticsphilosophy} between \textit{semantic} interpretation (i.e., meaning encoded directly in the utterance and does not change based on its external context), and \textit{pragmatic} interpretation (that depends on extra-linguistic information). The latter has gathered significant interest in the NLP community recently \citep{bender-koller-2020-climbing,bisk-etal-2020-experience}, focusing on language understanding, when grounded in an external context \citep{fried-etal-2023-pragmatics}. To a large extent, the focus of such studies has been on grounding language in a perceptual environment (e.g., image captioning \citep{andreas-klein-2016-reasoning,sharma-etal-2018-conceptual,alikhani-etal-2020-cross}, instruction following \citep{wang-etal-2016-learning-language,suhr-etal-2019-executing,lachmy-etal-2022-draw}, and game playing \citep{potts2012goal,udagawa2019natural} tasks). Unlike these works, in this paper, we focus on \textit{grounding language in a social context}, i.e., modeling the common ground \citep{clark1991grounding,traum1994computational,stalnaker2002common} between the author and their social media followers, that enables understanding an otherwise  highly ambiguous utterances.
The \textit{Social Context Understanding}, needed for building successful models for such tasks, can come from a wide variety of sources. The politician's affiliation and historical stances on the issue provide can capture crucial social context. Social relationships, knowledge about the involved entities, and related prior and upcoming events form important part of the puzzle as well. In \cref{fig:intro_example} event $\#1$, combining the event information (\textit{school shooting}) with the speakers' gun control stances, would facilitate understanding the \textit{intended meaning} of the text.

\textbf{The main motivation of this paper work is to operationalize the \textit{`Social Context Grounding'} problem as a pragmatic understanding task}. From a practical perspective, this would enable the creation of better NLP-CSS models that can process social media text in settings that require contextualized understanding. We suggest several  datasets, designed to evaluate this ability in computational models. These task capture the intended meaning at different level of granularity. At the most basic level, providing the social context can help identify the entities targeted, and the sentiment towards them. In \cref{fig:intro_example}, the social context \textlangle event\#1, Harris\textrangle  and the text \textit{``we need to keep our teachers safe''} $\Rightarrow$ \textit{``\underline{negative} attitude towards \underline{guns}''}. A more nuanced account of meaning, which we formulate as a separate task, captures the specific means in which the negative attitude is expressed (the Interpretation in \cref{fig:intro_example}). 
We additionally present two datasets corresponding to these tasks, namely, \textit{`Target Entity and Sentiment Detection'} and \textit{`Vague Text Disambiguation'}. In the first, the goal is to predict: $1$) whether a given entity is the \textit{intended target} of a politician's tweet and $2$) the sentiment towards the intended targets. We explicitly focus on tweets that \textit{do not always mention the targets} in their text to incentivize modeling the pragmatic communicative intent of the text. In the second task, given an ambiguous political message such as \textit{``We demand justice''} and its social context (\textit{associated event}, \& the \textit{author's party affiliation}), the task is to identify a \textit{plausible} unambiguous explanation of the message. Note that the ground truth for all these tasks is based on human pragmatic interpretation, i.e., ``\textit{guns}'' is a negative target of ``\textit{we need to keep our teachers safe}'', despite not being mentioned in the text, since it was perceived in this way by a team of human annotators reading the tweet and knowing social context.
We show examples of each task in \cref{tab:data_stats}. We describe the datasets in detail in \cref{sec:datasets}.

We evaluate the performance of various models, as a way to test the need for social context and compare different approaches for modeling it. These include pre-trained LM-based classifiers, and LLM in-context learning~\citep{gpt3,gpt-neox-20b}, which use a textual representation of the social context. We also adopt an existing graph-based discourse contextualization framework \citep{pujari-goldwasser-2021-understanding,feng2022political}, to explicitly model the social context needed to solve the proposed tasks. Our  results demonstrate that the discourse contextualization models outperform other models on both tasks. We present an error analysis to gain further insights. We describe the models in \cref{sec:baselines} and the results in \cref{sec:results}.

We also present a qualitative visualization of a political event, \textit{Brett Kavanaugh Supreme Court Nomination} (\cref{sec:event_visualization}), from target entity-sentiment perspective. It showcases a unique summary of the event discourse. We perform human evaluation on our \textit{`Vague Text Disambiguation'} dataset, and observe that humans find this task much easier than the evaluated models. We also present observations of human vs. LLM errors in disambiguation. In summary, our contributions are: 
\begin{enumerate}[noitemsep,nolistsep]
    \item Defining and operationalizing the ‘\textit{Social Context Grounding}’ task in political discourse
    \item Evaluating various state-of-the-art context representation models on the task. We adopt existing discourse contextualization framework for the proposed tasks, and evaluate GPT-3's in-context learning performance, as well.
    \item Performing human studies to benchmark the dataset difficulty and GPT-3 generation performance, when compared to human workers.\footnote{Our data and code is at \url{https://github.com/pujari-rajkumar/language-in-context}}
\end{enumerate}


\section{Related Work}\label{sec:related_work}
\noindent \textbf{Pragmatic Language Grounding}  gained significant focus recently \citep{bender-koller-2020-climbing,bisk-etal-2020-experience} following the rise of Pretrained Language Models \citep{devlin-etal-2019-bert,Liu2019RoBERTaAR,gpt3} as unified NLP models. Most grounding tasks address multi-modal or physical environment descriptions \citep{barnard2003matching,vogel2010learning,chen2011learning,tellex2011understanding,mitchell2012midge,anderson2018vision}. We refer the reader to \cite{fried-etal-2023-pragmatics} for a thorough overview. In contrast, we focus on grounding language in a \textit{social} context.

\noindent \textbf{Social Context Modeling} \citet{hovy-yang-2021-importance} show that modeling social context is necessary for human-level NLU. As political messages are often targeted at the voter base aware of the political context \citep{Weber2021AmplifyingIT,human_message_politics2011}, they are vague by design. Several previous works model social context for entity linking \citep{yang-etal-2016-toward}, social media connections relationship for fake news detection \cite{baly-etal-2018-predicting,mehta2022tackling} and, political bias detection \citep{li-goldwasser-2019-encoding,baly2020we}. These works model partial aspects of social context, relevant to their tasks. Two recent frameworks aim to capture social context holistically \citep{pujari-goldwasser-2021-understanding,feng2022political}. Evaluation tasks presented in both works 
show interesting social context understanding but are not fully representative of the challenges of \textit{Social Context Grounding}. \citet{zhan2023socialdial} propose a dataset for dialogue understanding addressing general social commonsense.

\noindent \textbf{Related Semantic and Pragmatic tasks} closest to our \textit{Target Entity Sentiment Identification} task is \textit{Stance Detection} in social media~\cite{DBLP:journals/corr/MohammadSK16,stance-detection-aldayel-2020}. To clarify our contribution, \citet{DBLP:journals/corr/MohammadSK16}, a popular SemEval task, looks at sentiment towards $5$ targets, while our data has $362$ unique targets. \citet{allaway-mckeown-2020-zero} and \citet{zhang-etal-2022-generative} also propose stance datasets on tweets. But, they focus mainly on semantic understanding of text that allows them to predict agreement or disagreement with well-defined statements.
Our \textit{Vague Text Disambiguation} task is related to recent works that study implicit inferences~\cite{hoyle-etal-2023-natural}, and pragmatic understanding \cite{hu-etal-2023-fine}. However, our tasks evaluate pragmatic understanding using an explicit context, absent in those tasks.

\section{Social Context Grounding Tasks}\label{sec:datasets}
We design and collect two datasets for \textit{Social Context Grounding} evaluation, and define three pragmatic interpretation tasks.
%
 In the \textit{Tweet Target Entity and Sentiment} dataset, we collect annotations of opinionated tweets from known politicians for their intended targets and sentiments towards them. We focus on {three} political events for this task. The dataset and its collection are described below in \cref{sec:target_entity_dataset}.
In the \textit{Vague Text Disambiguation Task}, we collect plausible explanations of vague texts, given the social context, consisting of \textit{author affiliation} and \textit{specific event}. We focus on {eight} political events. This dataset is detailed in \cref{sec:vague_text_dataset}. Examples and data statistics are shown in \cref{tab:data_stats}.

\subsection{Tweet Target Entity and Sentiment Task}\label{sec:target_entity_dataset}
In this task, given a tweet \textit{T}, its context, and an entity \textit{E}, the objective is to predict whether or not \textit{E} is a target of \textit{T} and the sentiment towards \textit{E}. Political discourse often contains opinionated discourse about world events and social issues. We collect tweets that don't directly mention the target entities. Thus, connecting the text with the event details and the author's general perspectives is necessary to solve this task effectively. We pick the focal entities for the given event and let human annotators expand on that initial set, based on their interpretation of the contextualized text. A \textit{target} entity is conceptualized as an entity present in the full intended interpretation of the tweet.

We focus our tweet collection on three recent divisive events: \textit{George Floyd Protests}, \textit{2021 US Capitol Attacks}, and \textit{Brett Kavanaugh's Supreme Court Nomination}. 
We identify relevant participating entities for each of the three events. Examples of the involved entities for the event \textit{George Floyd Protests} were \textit{George Floyd, United States Police, Derek Chauvin, Donald Trump, Joe Biden, United States Congress, Black people, Democratic Party, Republican Party, BLM, Antifa}.

\subsubsection{Target-Sentiment Data Collection}\label{sec:target_entity_collection}
We filter $3,454$ tweets for the \textit{three} events using hashtags, keyword-based querying, and the dates of the event-based filtering from the Congress Tweets repository corpus\footnote{\url{https://github.com/alexlitel/congresstweets}}. We collect a subset of $1,779$ tweets that contain media (images/video) to increase the chances of the tweet text not containing the target entity mentions. Then, we use $6$ in-house human annotators and Amazon Mechanical Turk (AMT) workers who are familiar with the event context for annotation. We ask them to annotate the targeted entities and sentiments towards the targets. The authors of this paper also participated in the annotation process. We provide them with entity options based on the event in the focus of the tweet. Annotators are allowed to add additional options if needed. We also ask the annotators to mark non-targets for each tweet. We instruct them to keep the non-targets as relevant to the event as possible to create harder negative examples. Each tweet is annotated by three annotators. We filter $865$ unique tweets with $5,891$ annotations, with majority agreement on each tweet. All the AMT annotations were additionally verified by in-house annotators for correctness. AMT workers were paid USD $1$ per tweet. It took $3$ minutes on average for each assignment, resulting in an hourly pay of USD $20$. We include screenshots of the collection task GUIs in the appendix. 
We split the train, and test sets by events, authors, and targets to incentivize testing the general social grounding capabilities of the models. The test set also consists of authors, targets, and events not seen in the training set. We use \textit{Capitol Riots} event for the test set of \textit{Target Entity and Sentiment} Task. We split the examples into $4,370$ train, $511$ development, and $1,009$ test examples. We compute the mean Cohen’s kappa score for annotations and report inter-annotator agreement for annotated targets (0.47) and sentiment (0.73)

\subsection{Vague Text Disambiguation Task}\label{sec:vague_text_dataset}

The task of \textit{Vague Text Disambiguation} is designed to capture pragmatic interpretation at a finer-grained level. It can be viewed as a variant of the well known paraphrase task, adapted for the social context settings. The model is evaluated on its ability to identify plausible interpretations (i.e., a sentence explicitly describing the author's intent) of an ambiguous quote given the event context and author's affiliation. 
E.g., ``\textit{protect our children from mass shootings}'' could easily be disambiguated as either ``\textit{ban guns}'' or ``\textit{arm teachers}'' when the author's stance on the issue of `\textit{gun rights}' is known. 

Our data collection effort is designed to capture different aspects of social context grounding and facilitate detailed error analysis. Defined as a binary classification task over tuples {\small \textlangle \texttt{Party, Event, Vague text, Explicit text}\textrangle}, we create negative examples by flipping tuple elements values of positive examples. This allows us to evaluate whether models can capture event relevance, political stance, or constrain the interpretation based on the vague text. For example, in the context of Event \#1 in \cref{fig:intro_example}, we can test if models simply capture the correlation between Democrats and negative stance towards guns access by replacing the vague text to \textit{``let your voice be heard''}, which would make the interpretation in \cref{fig:intro_example} implausible despite being consistent with that stance, while other consistent interpretations would be plausible (e.g.,  \textit{``go outside and join the march for our lives''}).

\subsubsection{Vague Text Data Collection}\label{sec:vague_text_collection}
Data collection was done in several steps. (1)\textbf{Vague Texts Collection}.  We collected vague text candidates from tweets by US politicians (i.e. senators and representatives) between the years $2019$ to $2021$ from Congress Tweets corpus. We identified a list of $9$ well-known events from that period and identified event-related tweets using their time frame and relevant hashtags. We used a pre-trained BERT-based \citep{devlin-etal-2019-bert} NER model to collect tweets that contain few or no entity mentions to identify potential candidates for vague texts. We manually identified examples that could have contrasting senses by flipping their social context. We obtain $93$ vague text candidates via this process.

\noindent(2) \textbf{In-Context Plausible Meaning Annotation}. We match the $93$ ambiguous tweets with different events that fit them. We use both Democrat/Republican as the author party affiliation. We obtain $600$ \textit{context-tweet} pairs for AMT annotation. For each tweet, we ask AMT workers to annotate the following two aspects: 1) sentiment towards the three most relevant entities in the event (sanity check) and 2) a detailed explanation of the \textit{intended meaning} given the event and author's party affiliation. We obtain $469$ reasonable annotations. After this step, each annotation was screened by in-house annotators. We ask three in-house annotators to vote on the \textit{correctness}, \textit{appropriateness}, and \textit{plausibility} of the annotation given the context. Thus, we create a total of $374$ examples.

\noindent(3)\textbf{LLM-based Data Expansion}. Using these examples, we further generate candidates for the task using LLM few-shot prompting. We use the examples from the previous step as in-context few-shot examples in the prompt. We use GPT-NeoX \citep{gpt-neox-20b} and GPT-3 \citep{gpt3} for candidate generation. Manual inspection by three in-house annotators is performed for each generated answer to ensure data quality. We generate $928$ candidates using GPT-NeoX and GPT-3. Manual filtering results in $650$ generations that pass the quality check. After removing redundant samples, we obtain $365$ additional examples. Thus, we obtain a total of $739$ annotations for this task. Then, for each of the $739$ examples, we ask in-house annotators to select $3$ relevant negative options from the pool of explanations. We instruct them to pick hard examples that potentially contain overlapping entities with the gold answer. This results in $2,956$ binary classification data samples. We analyze and discuss the results of human validation of large LM generations in \cref{sec:analysis}).

This process allows us to create three variants of the task: binary-classification, multiple-choice and generation variants. We evaluate several classification models on the binary classification variant (Tab.\ref{tab:para_results}). We evaluate LLMs on the generation variant (\S\ref{sec:llm_quality_analysis}). We benchmark humans and the best models on the multiple-choice variant (\S\ref{sec:human_perf}).

Similar to the previous task, we split the train, test sets by events, and vague text to test the general social understanding capabilities of the model. We reserve \textit{Donald Trump's second impeachment verdict} event for the test set. We also reserve Democratic examples of $2$ events and Republican examples of $2$ events exclusively for the test set. We split the dataset into $1,916$ train, $460$ development, and $580$ test examples. $180$ of the test examples are from events/party contexts unseen in train data.

\section{Modeling Social Context}\label{sec:baselines}
\begin{table*}[thb]
\centering
\resizebox{400pt}{!}{
\begin{tabular}{clcccc|cccc}
\hline
\multicolumn{2}{c}{\multirow{2}{*}{Model}}                                                                                                        & \multicolumn{4}{c|}{\textbf{Target Identification}}               & \multicolumn{4}{c}{\textbf{Sentiment Identification}}             \\ \cline{3-10} 
\multicolumn{2}{c}{}                                                                                                                              & Prec           & Rec            & Macro-F1       & Acc            & Prec           & Rec            & Macro-F1       & Acc            \\ \hline
\multirow{2}{*}{\begin{tabular}[c]{@{}c@{}}No Context\\ Baselines\end{tabular}}                    & \multicolumn{1}{l|}{BERT-large}              & 69.09          & 72.35          & 68.83          & 70.56          & 58.74          & 60.17          & 58.95          & 58.37          \\
                                                                                                   & \multicolumn{1}{l|}{RoBERTa-base}            & 66.58          & 69.54          & 65.14          & 66.40          & 61.68          & 61.27          & 61.36          & 60.65          \\ \hline
\multirow{2}{*}{\begin{tabular}[c]{@{}c@{}}PLMs\\ +Twitter Bio Context\end{tabular}}               & \multicolumn{1}{l|}{BERT-large + user-bio}   & 69.03          & 71.86          & 69.34          & 71.66          & 60.02          & 60.44          & 60.13          & 59.86          \\
                                                                                                   & \multicolumn{1}{l|}{RoBERTa-base + user-bio} & 65.83          & 68.65          & 64.79          & 66.30          & 60.06          & 59.91          & 59.94          & 59.46          \\ \hline
\multirow{2}{*}{\begin{tabular}[c]{@{}c@{}}PLMs\\ +Wikipedia Context\end{tabular}}                 & \multicolumn{1}{l|}{BERT-large + wiki}       & 63.58          & 65.78          & 60.33          & 61.05          & 53.48          & 56.44          & 53.9           & 53.32          \\
                                                                                                   & \multicolumn{1}{l|}{RoBERTa-base + wiki}     & 69.02          & 72.32          & 68.62          & 70.27          & 57.62          & 59.10          & 58.07          & 58.28          \\ \hline
\multirow{2}{*}{LLMs}                                                                              & \multicolumn{1}{l|}{GPT-3 0-shot}            & 69.25          & 70.58          & 69.77          & 73.78          & 56.20          & 55.04          & 54.18          & 56.80          \\
                                                                                                   & \multicolumn{1}{l|}{GPT-3 4-shot}            & 69.81          & 72.99          & 66.45          & 67.03          & 58.12          & 57.10          & 55.00          & 57.51          \\ \hline
\multirow{4}{*}{\begin{tabular}[c]{@{}c@{}}Static Contextutalized\\ Embedding Models\end{tabular}} & \multicolumn{1}{l|}{RoBERTa-base + PAR Embs} & 68.38          & 71.63          & 67.67          & 69.18          & 55.01          & 56.89          & 55.51          & 55.40          \\
                                                                                                   & \multicolumn{1}{l|}{BERT-large + PAR Embs}   & 65.40          & 67.33          & 60.25          & 60.56          & 55.24          & 57.54          & 55.89          & 55.80          \\
                                                                                                   & \multicolumn{1}{l|}{RoBERTa-base + DCF Embs} & \textbf{72.89} & \textbf{75.95} & \textbf{73.56} & \textbf{75.82} & 63.05          & 63.52          & 62.90          & 63.03          \\
                                                                                                   & \multicolumn{1}{l|}{BERT-large + DCF Embs}   & 68.76          & 72.02          & 68.32          & 69.97          & 61.59          & 63.25          & 61.22          & 60.75          \\ \hline
\multirow{2}{*}{\begin{tabular}[c]{@{}c@{}}Discourse\\ Contextualized Models\end{tabular}}         & \multicolumn{1}{l|}{BERT-large + DCF}        & 71.12          & 74.61          & 71.17          & 72.94          & \textbf{65.81} & \textbf{65.25} & \textbf{65.34} & \textbf{65.31} \\
                                                                                                   & \multicolumn{1}{l|}{RoBERTa-base + DCF}      & 70.44          & 73.86          & 70.39          & 72.15          & 63.45          & 63.34          & 63.37          & 63.23          \\ \hline
\end{tabular}
}
\caption{Results of baseline experiments on \textit{Target Entity} (binary task) and \textit{Sentiment} ($4$-classes) test sets. 
We report macro-averaged Precision, macro-averaged Recall, macro-averaged F1, and Accuracy metrics. }
\label{tab:targ_sent_results}
\end{table*}



\begin{table}[htb]
\resizebox{200pt}{!}{
\begin{tabular}{lcccc}
\hline
\multicolumn{1}{c}{\multirow{2}{*}{\textbf{Model}}} & \multicolumn{4}{c}{\textbf{Vague Text Disambiguation}}                                                                                          \\ \cline{2-5} 
\multicolumn{1}{c}{}                                & \multicolumn{1}{c}{\textbf{Prec}} & \multicolumn{1}{c}{\textbf{Rec}} & \multicolumn{1}{c}{\textbf{Macro-F1}} & \multicolumn{1}{c}{\textbf{Acc}} \\ \hline
\multicolumn{5}{l}{\textbf{No Context Baselines}}                                                                                                                                               \\ \hline
\multicolumn{1}{l|}{BERT-large}                     & $52.24$ & $55.58$ & $50.28$ & $53.75$\\
\multicolumn{1}{l|}{RoBERTa-base}                   & $55.3$ & $51.82$ & $54.53$ & $56.08$ \\ \hline

\multicolumn{5}{l}{\textbf{PLMs + Wikipedia Context}}                                                                                                                                         \\ \hline
\multicolumn{1}{l|}{BERT-large + wiki}              &   $52.31$ & $46.90$ & $66.87$ & $76.03$  \\
\multicolumn{1}{l|}{BERT-base + wiki}            &   $51.85$ & $38.62$ & $64.36$ & $75.69$ \\ \hline
\multicolumn{5}{l}{\textbf{LLMs}}                                                                                                                                                             \\ \hline
\multicolumn{1}{l|}{GPT-3 0-shot}                   &   \textbf{63.10} & $62.92$ & $62.58$ & $63.5$ \\
\multicolumn{1}{l|}{GPT-3 4-shot}                   &   $62.05$ & $62.29$ & $61.86$ & $62.04$ \\ \hline
\multicolumn{5}{l}{\textbf{Static Contextutalized Embedding Models}}                                                                                                                                        \\ \hline
\multicolumn{1}{l|}{BERT-large + PAR}               & $47.68$ & $49.66$ & $65.53$ & $73.79$  \\
\multicolumn{1}{l|}{BERT-base + PAR}             &   $45.93$ & $54.48$ & $65.49$ & $72.59$ \\
\multicolumn{1}{l|}{BERT-large + DCF Embs}           &  $47.18$ & \textbf{63.45} & $67.55$ & $73.10$ \\
\multicolumn{1}{l|}{BERT-base + DCF Embs}        &   $56.58$ & $59.31$ & \textbf{71.71} & \textbf{78.45} \\ \hline
\multicolumn{5}{l}{\textbf{Discourse Contextualization Models}}                                                                                                                                        \\ \hline
\multicolumn{1}{l|}{BERT-large + DCF}               &   $52.76$ & $59.31$ & $69.94$ & $76.55$  \\
\multicolumn{1}{l|}{BERT-base + DCF}             &     $52.73$ & $60.00$ & $70.06$ & $76.55$ \\ \hline
\end{tabular}
}
\caption{Results of baseline experiments on \textit{Vague Text Disambiguation} dataset test split, a binary classification task. We report macro-averaged Precision, macro-averaged Recall, macro-averaged F1, and Acc. metrics}
\label{tab:para_results}
\end{table}

The key technical question this paper puts forward is how to model the social context, such that the above tasks can be solved with high accuracy. We observe that humans can perform this task well (\cref{sec:human_perf}), and evaluate different context modeling approaches in terms of their ability to replicate human judgments. These correspond to \texttt{No Context}, \texttt{Text-based} context representation (e.g., Twitter Bio, relevant Wikipedia articles), and \texttt{Graph-based} context representation, simulating the social media information that human users are exposed to when reading the vague texts.

We report the results of all our baseline experiments in \cref{tab:targ_sent_results} and \cref{tab:para_results}. 
The first set of results evaluate fine-tuned pre-trained language models (PLM), namely BERT \citep{devlin-etal-2019-bert} and RoBERTa \citep{Liu2019RoBERTaAR}, with three stages of modeling context. Firstly, we evaluate no contextual information setting. Second, we include the authors' Twitter bios as context. Finally, we evaluate the information from the author, event, and target entity Wikipedia pages as context (models denoted \texttt{PLM Baselines \{No,  Twitter Bio, Wikipedia\} Context}, respectively).

We evaluate GPT-3\footnote{gpt-3.5-turbo-1106 via OpenAI API} in \textit{zero-shot} and \textit{four-shot} in-context learning paradigm on both tasks. We provide contextual information in the prompt as short event descriptions and authors' affiliation descriptions. Note that GPT-3 is trained on news data until Sep. $2021$ which includes the events in our data (models denoted \texttt{LLM Baseline}).

 We evaluate the performance of politician embeddings from Political Actor Representation (PAR) \citep{feng2022political} and Discourse Contextualization Framework (DCF) \citep{pujari-goldwasser-2021-understanding} models.
 (models denoted \texttt{Static Contextutalized Embeddings}). We use PAR embeddings available on their GitHub repository\footnote{\url{https://github.com/BunsenFeng/PAR}}. For DCF model, we use released pre-trained models from GitHub repository\footnote{\url{https://github.com/pujari-rajkumar/compositional_learner}} to generate author, event, text, and target entity embeddings. We evaluate the embeddings on both tasks. We briefly review these models in \cref{sec:dcf} \& \cref{sec:par}. 

Finally, we use tweets of politicians from related previous events and build context graphs for each data example as proposed in \citet{pujari-goldwasser-2021-understanding}. We use Wikipedia pages of authors, events, and target entities to add social context information to the graph. Then, we train the Discourse Contextualization Framework (DCF) for each task and evaluate its performance on both tasks (models denoted \texttt{Discourse Contextualization Model}). Further details of our baseline experiments are presented in subsection \cref{sec:experiments}. Results of our baseline experiments are discussed in \cref{sec:results}.

\subsection{Discourse Contextualization Framework} \label{sec:dcf}
Discourse Contextualization Framework (DCF) \citep{pujari-goldwasser-2021-understanding} leverages relations among social context components to learn contextualized representations for text, politicians, events, and issues. It consists of \textit{encoder} and \textit{composer} modules that compute holistic representations of the context graph. The encoder creates an initial representation of nodes. Composer propagates the information within the graph to update node representations. They define link prediction learning tasks over context graphs to train the model. They show that their representations significantly outperform several PLM-based baselines trained using the same learning tasks.

\subsection{Political Actor Representation} \label{sec:par}
\citet{feng2022political} propose the \textit{Political Actor Representation} (PAR) framework, a graph-based approach to learn more effective politician embeddings. They propose three learning tasks, namely, $1$) Expert Knowledge Alignment $2$) Stance Consistency training \& $3$) Echo chamber simulation, to infuse social context into the politician representations. They show that PAR representations outperform SOTA models on \textit{Roll Call Vote Prediction} and \textit{Political Perspective Detection}. 

\subsection{Experimental Setup} \label{sec:experiments}

\textit{Target Entity Detection} is binary classification with \textlangle \textit{author, event, tweet, target-entity}\textrangle as input and \textit{target/non-target} label as output. \textit{Sentiment Detection} is set up as $4$-way classification. Input is the same as the target task and output is one of: \{\textit{positive, neutral, negative, non-target}\}. \textit{Vague Text Disambiguation} is a binary classification task with \textlangle\textit{party-affiliation, event, vague-text, explanation-text}\textrangle and a \textit{match/no-match} label as output.

In phase 1 no-context baselines, we use the author, event, tweet, and target embeddings generated by PLMs. We concatenate them for input. In Twitter-bio models, we use the author's Twitter bio embeddings to represent them. Wiki context models receive Wikipedia page embeddings of author, event, and target embeddings. \textit{It is interesting to note that the Wikipedia context models get all the information needed to solve the tasks.}. In phase 2 LLM experiments, we use train samples as in-context demonstrations. We provide task and event descriptions in the prompt. In phase 3 PAR models, we use politician embeddings released on the PAR GitHub repository to represent authors. We replace missing authors with their wiki embeddings. For the \textit{Vague Text} task, we average PAR embeddings for all politicians of the party to obtain party embeddings. For DCF embedding models, we generate representations for all the inputs using context graphs. We also use authors' tweets from relevant past events. We build graphs using author, event, tweet, relevant tweets, and target entity as nodes and edges as defined in the original DCF paper. In phase 4, we use the same setup as the DCF embedding model and additionally back-propagate to DCF parameters. This allows us to fine-tune the DCF context graph representation for our tasks.

\section{Results} \label{sec:results}
The results of our baseline experiments are described in Tab. \ref{tab:targ_sent_results} and \ref{tab:para_results}. We evaluate our models using macro-averaged precision, recall, F1, and accuracy metrics (due to class imbalance, we focus on macro-F1). Several patterns, consistent across all tasks, emerge. \textbf{First}, \textit{modeling social context is still an open problem}. None of our models were able to perform close to human level. \textbf{Second}, \textit{adding context can help performance}, compared to the No-Context baselines,  models incorporating context performed better, with very few exceptions. \textbf{Third}, \textit{LLMs are not the panacea for social-context pragmatic tasks}. Despite having access to a textual context representation as part of the prompt, and having access to relevant event-related documents during their training phase, these models under-perform compared to much simpler models that were fine-tuned for this task. \textbf{Finally}, \textit{explicit context modeling using the DCF model consistently leads to the best performance}. 
The DCF model mainly represents the social context in the form of text documents for all nodes. Further symbolic addition of other types of context such as social relationships among politicians and relationships between various nodes could further help in achieving better performance on these tasks.
In the \textit{Target Entity} task, RoBERTa-base + DCF embeddings obtain $73.56$ F1 vs. $68.83$ for the best no-context baseline. Twitter bio and wiki-context hardly improve, demonstrating the effectiveness of modeling contextual information explicitly vs. concatenating context as text documents. No context performance well above the random performance of $50$ F1 indicates the bias in the target entity distribution among classes. We discuss this in \cref{sec:event_visualization}. In \textit{Sentiment Identification} task, we see that BERT-large + DCF back-propagation outperforms all other models. \textit{Vague Text Disambiguation} task results in \cref{tab:para_results} show that DCF models outperform other models significantly. $71.71$ F1 is obtained by BERT-base + DCF embeddings. BERT-base performing better than bigger PLMs might be due to DCF model's learning tasks being trained using BERT-base embeddings.

\section{Analysis and Discussion}\label{sec:analysis}
\begin{table*}[htb]
\centering
\resizebox{400pt}{!}{
\begin{tabular}{|cc|ccccc|cc|}
\hline
\multicolumn{2}{|c|}{\textbf{Democrat Only Entities}}                                                                                                                                                       & \multicolumn{5}{c|}{\textbf{Common Entities}}                                                                                                                                                                                                                                                                                                                                                                                                                                                                                                                                                                                                            & \multicolumn{2}{c|}{\textbf{Republican Only Entities}}                                                                                                                                                                                      \\ \hline
\multicolumn{1}{|c|}{\multirow{2}{*}{\textbf{Target}}}                                                                  & \multirow{2}{*}{\textbf{Sentiment}}                                               & \multicolumn{2}{c|}{\textbf{Agreed-Upon Entities}}                                                                                                                                                                                            & \multicolumn{3}{c|}{\textbf{Divisive Entities}}                                                                                                                                                                                                                                                                                                                                                                     & \multicolumn{1}{c|}{\multirow{2}{*}{\textbf{Target}}}                                                                                         & \multirow{2}{*}{\textbf{Sentiment}}                                                         \\ \cline{3-7}
\multicolumn{1}{|c|}{}                                                                                                  &                                                                                   & \multicolumn{1}{c|}{\textbf{Target}}                                                                                          & \multicolumn{1}{c|}{\textbf{Sentiment}}                                                            & \multicolumn{1}{c|}{\textbf{Sentiment (D)}}                                                                             & \multicolumn{1}{c|}{\textbf{Target}}                                                                                                                                       & \textbf{Sentiment (R)}                                                                           & \multicolumn{1}{c|}{}                                                                                                                         &                                                                                             \\ \hline
\multicolumn{1}{|c|}{\begin{tabular}[c]{@{}c@{}}Anita Hill\\ Patty Murray \\ Merrick Garland\\ Jeff Flake\end{tabular}} & \begin{tabular}[c]{@{}c@{}}Positive\\ Positive\\ Positive\\ Negative\end{tabular} & \multicolumn{1}{c|}{\begin{tabular}[c]{@{}c@{}}US Supreme Court\\ US Senate \\ FBI\\ Judiciary Committee\end{tabular}} & \multicolumn{1}{c|}{\begin{tabular}[c]{@{}c@{}}Neutral\\ Neutral\\ Neutral\\ Neutral\end{tabular}} & \multicolumn{1}{c|}{\begin{tabular}[c]{@{}c@{}}Positive\\ Positive\\ Positive\\ Negative\\ Negative\\ Negative\end{tabular}} & \multicolumn{1}{c|}{\begin{tabular}[c]{@{}c@{}}Christine Blasey Ford \\ Deborah Ramirez\\ Julie Swetnick\\ Brett Kavanaugh \\ Donald Trump\\ Mitch McConnell\end{tabular}} & \begin{tabular}[c]{@{}c@{}}Negative\\ Negative\\ Negative\\ Positive\\ Positive\\ Positive\end{tabular} & \multicolumn{1}{c|}{\begin{tabular}[c]{@{}c@{}}Susan Collins \\ Chuck Grassley\\ Diane Feinstein\\ Chuck Schumer\\ Sean Hannity\end{tabular}} & \begin{tabular}[c]{@{}c@{}}Positive\\ Positive\\ Negative\\ Negative\\ Neutral\end{tabular} \\ \hline
\end{tabular}
}
\caption{Target Entity-Sentiment centric view of \textit{Kavanaugh Supreme Court Nomination} discourse}
\label{tab:target_visualization}
\end{table*}

\subsection{Ablation Analysis on Vague Text Task}
We report ablation studies in \cref{tab:vague_text_ablation} on the Vague Text task test set. We consider $5$ splits: \textbf{(1)} Unseen Party: \textlangle\textit{party, event}\textrangle not in the train set but \textlangle\textit{opposing-party, event}\textrangle is present, \textbf{(2)} Unseen Event: \textlangle \textit{party} \textrangle not in train set, \textbf{(3)} Flip Event: negative samples with corresponding `event flipped-party/vague tweet matched' positive samples in train set and analogous \textbf{(4)} Flip Party and \textbf{(5)} Flip Tweet splits. We observe the best model in each category. They obtain weaker performance on unseen splits, as expected, unseen events being the hardest. Contextualized models achieve higher margins. DCF gains 7.6(13.2\%) and DCF embeddings attain 8.12(20.42\%) macro-F1 improvement over BERT-base+wiki compared to respective margins of 8.86\% and 11.42\% on the full test set. In the flip splits with only negative examples, accuracy gain over random baseline for all splits is seen. This indicates that models learn to jointly condition on context information rather than learn spurious correlations over particular aspects of the context. Specifically, flip-tweet split results indicate that models don't just learn party-explanation mapping.

\begin{table}[htb]
    \centering
    \resizebox{200pt}{!}{
       \begin{tabular}{lc|c|c|c|c}
        \hline
        \multicolumn{1}{c}{\multirow{2}{*}{\textbf{Data Split}}}                           & \textbf{\begin{tabular}[c]{@{}c@{}}Unseen\\ Party\end{tabular}} & \textbf{\begin{tabular}[c]{@{}c@{}}Unseen\\ Event\end{tabular}} & \textbf{\begin{tabular}[c]{@{}c@{}}Flip\\ Tweet\end{tabular}} & \textbf{\begin{tabular}[c]{@{}c@{}}Flip\\ Event\end{tabular}} & \textbf{\begin{tabular}[c]{@{}c@{}}Flip\\ Party\end{tabular}} \\ \cline{2-6} 
        \multicolumn{1}{c}{}                                                               & \textbf{Ma-F1}                                                  & \textbf{Ma-F1}                                                  & \textbf{Acc}                                                  & \textbf{Acc}                                                  & \textbf{Acc}                                                  \\ \hline
        \multicolumn{1}{l|}{Random}                                                        & 44.70                                                           & 29.69                                                           & 75                                                            & 75                                                            & 75                                                            \\
        \multicolumn{1}{l|}{BERT-base+wiki}                                                & 57.58                                                           & 39.76                                                           & 88.14                                                         & 89.77                                                         & 87.77                                                         \\
        \multicolumn{1}{l|}{\begin{tabular}[c]{@{}l@{}}BERT-base\\ +DCF Embs\end{tabular}} & 61.79                                                           & 47.88                                                           & 86.10                                                         & 93.18                                                         & 84.57                                                         \\
        \multicolumn{1}{l|}{BERT-base+DCF}                                                 & 65.18                                                           & 45.65                                                           & 82.03                                                         & 89.77                                                         & 84.04                                                         \\ \hline
        \end{tabular}
    }
    \caption{Ablation Study Results on Vague Text Task}
    \label{tab:vague_text_ablation}
\end{table}

\subsection{Vague Text LLM Generation Quality} \label{sec:llm_quality_analysis}
We look into the quality of our LLM-generated disambiguation texts. While GPT-NeoX \citep{gpt-neox-20b} produced only $98$ good examples out of the $498$ generated instances with the rest being redundant, GPT-3 \citep{gpt3} performed much better. Among the $430$ generated instances, $315$ were annotated as good which converts to an acceptance rate of $20.04\%$ for GPT-NeoX and $73.26\%$ for GPT-3 respectively. In-house annotators evaluated the quality of the generated responses for how well they aligned with the contextual information. They rejected examples that were either too vague, align with the wrong ideology, or were irrelevant. In the prompt, we condition the input examples in all the few shots to the same event and affiliation as the input vague text. In comparison, the validation of AMT annotations for the same task yielded $79.8\%$ good examples even after extensive training and qualification tests. Most of the rejections from AMT were attributed to careless annotations.

\subsection{Vague Text Human Performance} \label{sec:human_perf}
We look into how humans perform on the \textit{Vague Text Disambiguation} task. We randomly sample $97$ questions and ask annotators to answer them as multiple-choice questions. Each vague text-context pair was given $4$ choices out of which only one was correct. We provide a brief event description along with all the metadata available to the annotator. Each question was answered by $3$ annotators. Among the $97$ answered questions, the accuracy was $94.85\%$, which shows this task is easy for humans who understand the context. Respective performance of best models on this subset of data for BERT-base+wiki ($54.89\%$), BERT-base+DCF-embs ($63.38\%$), BERT-base+DCF ($64.79\%$) is much lower than human performance.

\subsection{Target Entity Visualization} \label{sec:event_visualization}
The main goal of this analysis is to demonstrate the usefulness and inspire modeling research in the direction of entity-sentiment-centric view of political events. \cref{tab:target_visualization} visualizes one component of how partisan discourse is structured in these events.
We study \textit{Kavanaugh Supreme Court Nomination}. We identify discussed entities and separate them into divisive and agreed-upon entities. This analysis paints an accurate picture of the discussed event. We observe that the main entities of Trump, Dr. Ford, Kavanaugh, Sen. McConnell, and other accusers/survivors emerge as divisive entities. Entities such as Susan Collins and Anita Hill who were vocal mouthpieces of the respective party stances but didn't directly participate in the event emerge as partisan entities. Supreme Court, FBI, and other entities occur but only as neutral entities.

\subsection{DCF Context Understanding}
We look into examples that are incorrectly predicted using Wikipedia pages but correctly predicted by the DCF model in the appendix (\cref{tab:tweet_ens_analysis}). In examples $1$ \& $2$ of \textit{Target Entity-Sentiment} task, when the entity is not explicitly mentioned in the tweet, the Wiki-Context model fails to identify them as the targets. We posit that while the Wikipedia page of each relevant event will contain these names, explicit modeling of entities in the DCF model allows correct classification. Examples $1-3$ of \textit{Vague Text Disambiguation} task show that when no clear terms indicate the sentiment towards a view, the Wiki-Context model fails to disambiguate the tweet text. Explicit modeling of politician nodes seems to help the DCF model.

\section{Conclusion and Future Work}\label{sec:conclusion}
In this paper, we motivate, define, and operationalize `\textit{Social Context Grounding}' for political text. We build two novel datasets to evaluate social context grounding in NLP models that `are easy for humans' when the relevant social context is provided. We experiment with many types of contextual models. We show that explicit modeling of social context outperforms other models while lacking behind humans.

\section*{Acknowledgements}\label{sec:ackowledgements}
We thank Shamik Roy, Nikhil Mehta, and the anonymous reviewers for their vital feedback. The project was funded by NSF CAREER award IIS-2048001 and the DARPA CCU Program. The contents are those of the author(s) and do not necessarily represent the official views of, nor an endorsement by, DARPA, or the US Government.

\section*{Limitations}\label{sec:limitations}
Our work only addresses English language text in US political domain. We also build upon large language models and large PLMs which are trained upon huge amounts of uncurated data. Although we employed human validation at each stage, biases could creep into the datasets. We also don't account for the completeness of our datasets as it is a pioneering work on a new problem. Social context is vast and could have a myriad of components. We only take a step in the direction of social context grounding in this work. The performance on these datasets might not indicate full social context understanding but they should help in sparking research in the direction of models that explicitly model such context. Although we tuned our prompts a lot, better prompts and evolving models might produce better results on the LLM baselines. Our qualitative analysis is predicated on a handful of examples. They are attempts to interpret the results of large neural models and hence don't carry as much confidence as our empirical observations. 
We believe the insights from our findings will encourage more research in this area. For example, the development of discourse contextualized models that aim to model human-style understanding of background knowledge, emotional intelligence, and societal context understanding is a natural next step of our research.

\section*{Ethics Statement}\label{sec:ethics}
In this work, our data collection process consists of using both AMT and GPT-3. For the \textit{Target Entity and Sentiment} task, we pay AMT workers $\$1$ per HIT and expect an average work time of $3$ minutes. This translates to an hourly rate of $\$20$ which is above the federal minimum wage. For the \textit{Vague Text Disambiguation} task, we pay AMT workers $\$1.10$ per HIT and expect an average work time of $3$ minutes. This translated to an hourly rate of $\$22$.

We recognize collecting political views from AMT and GPT-3 may come with bias or explicit results and employ expert gatekeepers to filter out unqualified workers and remove explicit results from the dataset. Domain experts used for annotation are chosen to ensure that they are fully familiar with the events in focus. Domain experts were provided with the context related to the events via their Wikipedia pages, background on the general issue in focus, fully contextualized quotes, and authors’ historical discourse obtained from ontheissues.org. We have an annotation quid-pro-quo system in our lab which allows us to have a network of in-house annotators. In-house domain experts are researchers in the CSS area with familiarity with a range of issues and stances in the US political scene. They are given the information necessary to understand the events in focus in the form of Wikipedia articles, quotes from the politicians in focus obtained from ontheissues.org, and news articles related to the event. We make the annotation process as unambiguous as possible. In our annotation exercise, we ask the annotators to mark only high-confidence annotations that can be clearly explained. We use a majority vote from $3$ annotators to validate the annotations for the target entity task.

Our task is aimed at understanding and grounding polarized text in its intended meaning. We take examples where the intended meaning is clearly backed by several existing real-world quotes. We do not manufacture the meaning to the vague statements, we only write down unambiguous explanations where context clearly dictates the provided meaning. Applications of our research as we envision would be adding necessary context to short texts by being able to identify past discourse from the authors that are relevant to the particular text in its context. It would also be able to ground the text in news articles that expand upon the short texts to provide full context. 


\appendix

\section{GPT Prompts}
\textbf{Prompts for Target-Entity Task:}
\begin{tcolorbox}[colback=gray!10, colframe=gray!50, width=\columnwidth, boxrule=0.5mm, arc=2mm, auto outer arc, fontupper=\scriptsize]
\texttt{ 
\textbf{Event:} \textcolor{red}{<event>}\\
\textbf{Event background:} \textcolor{red}{<background-description>}\\
\textbf{Tweet:} \textcolor{red}{<tweet-text>}\\
\textbf{Author:} \textcolor{red}{<author-name>}\\
\textbf{Author Party:} \textcolor{red}{<party-affiliation>}\\
\textbf{Author background:} \textcolor{red}{<first two sentences of author-wiki-page>}\\
\textbf{Target Entity:} \textcolor{red}{<entity-name>}\\
\textbf{Entity background:} \textcolor{red}{<first two sentences of entity-wiki-page>}\\
\textbf{Task:} Identify if the given entity is a target of the tweet. A target entity is defined as an entity that would be present in the full unambiguous explanation of the tweet.\\
\textbf{Is the given entity a target entity of the tweet?} Answer yes or no.
}
\end{tcolorbox}

\noindent\textbf{Prompts for Target-Sentiment Task:}
\begin{tcolorbox}[colback=gray!10, colframe=gray!50, width=\columnwidth, boxrule=0.5mm, arc=2mm, auto outer arc, fontupper=\scriptsize]
\texttt{
\textbf{Event:} \textcolor{red}{<event>}\\
\textbf{Event background:} \textcolor{red}{<background-description>}\\
\textbf{Tweet:} \textcolor{red}{<tweet-text>}\\
\textbf{Author:} \textcolor{red}{<author-name>}\\
\textbf{Author Party:} \textcolor{red}{<party-affiliation>}\\
\textbf{Author background:} \textcolor{red}{<first two sentences of author-wiki-page>}\\
\textbf{Target Entity:} \textcolor{red}{<entity-name>}\\
\textbf{Entity background:} \textcolor{red}{<first two sentences of entity-wiki-page>}\\
\textbf{Task:} Identify the sentiment of the tweet towards the given target entity. Consider that the tweet is ambiguous and the entity might be implied without being explicitly mentioned.\\
\textbf{What is the sentiment of the tweet towards the target entity?} Answer with positive, negative, or neutral.
}
\end{tcolorbox}

\noindent \textbf{Prompts for Vague Text Task:}
\begin{tcolorbox}[colback=gray!10, colframe=gray!50, width=\columnwidth, boxrule=0.5mm, arc=2mm, auto outer arc, fontupper=\scriptsize]
\texttt{
\textbf{Event:} \textcolor{red}{<event>}\\
\textbf{Event background:} \textcolor{red}{<background-description>}\\
\textbf{Vague message:} \textcolor{red}{<vague-text>}\\
\textbf{Author Party:} \textcolor{red}{<party-affiliation>}\\
\textbf{Author background:} \textcolor{red}{<first two sentences of party-wiki-page>}\\
\textbf{Task:} Given the event, vague message, and party affiliation of the author, explain unambiguously the intended meaning of the vague message.\\
\textbf{Generate an unambiguous explanation for the vague message given the party affiliation of the author and the event in context.}
}
\end{tcolorbox}

\section{Reproducibility}
We use the HuggingFace Transformers \citep{wolf-etal-2020-transformers} library for PLMs. We use GPT-NeoX implementation by ElutherAI \citep{gpt-neox-20b} and GPT-3 \citep{DBLP:journals/corr/abs-2005-14165} via OpenAI API for our LLM baselines. We run $100$ epochs for all experiments. We use $10$ NVIDIA GeForce 1080i GPUs for our experiments. We use the train, development, and test splits detailed in \cref{sec:datasets} for our experiments. We use the development macro-F1 for early stopping. We run all our experiments using random seeds to ensure reproducibility. We experiment with a random seed value set to \{$13$\}. We set CUBLAS environment variables for reproducibility. All our code, datasets, and result logs are released publicly.

\section{Error Analysis}
\begin{table*}[tbh]
\resizebox{\textwidth}{!}{

\begin{tabular}{l|l}
\hline
\multicolumn{1}{c|}{\textbf{Target Entity and Sentiment Task}}                                                                                                                                                                                                                                                                                                                                                        & \multicolumn{1}{c}{\textbf{Vague Text Disambiguation Task}}                                                                                                                                                                                                                                                                                                                        \\ \hline
\begin{tabular}[c]{@{}l@{}}\textbf{Tweet}: Republicans held Justice Scalia’s seat open for more \\ than 400 days. Justice Kennedy’s seat has been vacant for \\ less than two months. It's more important to investigate a \\ serious allegation of sexual assault than to rush Kavanaugh \\ onto the Supreme Court for a lifetime appointment.\end{tabular}                                                           & \textbf{Tweet}: Thanks for this.                                                                                                                                                                                                                                                                                                                                    \\
\textbf{Author}: Adam Schiff (Democrat)                                                                                                                                                                                                                                                                                                                                                                                        & \textbf{Affiliation}: Democrat                                                                                                                                                                                                                                                                                                                                      \\
\textbf{Event}: Brett Kavanaugh Supreme Court nomination                                                                                                                                                                                                                                                                                                                                                               & \textbf{Event}: United States withdrawal from the Paris Agreement                                                                                                                                                                                                                                                                                                   \\
\textbf{Entity}: Christine Blasey Ford                                                                                                                                                                                                                                                                                                                                                                                 & \begin{tabular}[c]{@{}l@{}}\textbf{Paraphrase}: There's nothing surprising in withdrawing from\\ the Paris agreement. Thanks for not caring our environment and \\future generations.\end{tabular}                                                                                                                                                                \\
\textbf{Wiki-Context Prediction}: Not Target | \textbf{DCF Prediction}: Target (correct)                                                                                                                                                                                                                                                                                                                                                                 & \textbf{Wiki-Context Prediction}: No | \textbf{DCF Prediction}: Yes (correct)                                                                                                                                                                                                                                                                                                            \\\hline & \\
\begin{tabular}[c]{@{}l@{}}\textbf{Tweet}: We will not be intimidated. Democracy will not be \\ intimidated. We must hold the individuals responsible for the \\ Jan. 6th attack on the U.S. Capitol responsible.  Thank you \\ @RepAOC for tonight’s Special Order Hour and we will \\ continue our efforts to \#HoldThemAllAccountable.\end{tabular}                                                                 & \textbf{Tweet}: Let us say enough. Enough.                                                                                                                                                                                                                                                                                                                          \\
\textbf{Author}: Adriano Espaillat (Democrat)                                                                                                                                                                                                                                                                                                                                                                                         & \textbf{Affiliation}: Democrat                                                                                                                                                                                                                                                                                                                                      \\
\textbf{Event}: January 6 United States Capitol attack                                                                                                                                                                                                                                                                                                                                                                 & \begin{tabular}[c]{@{}l@{}}\textbf{Event}: Second impeachment of Donald Trump ended with not \\ guilty\end{tabular}                                                                                                                                                                                                                                                 \\
\textbf{Entity}: Donald Trump                                                                                                                                                                                                                                                                                                                                                                                          & \begin{tabular}[c]{@{}l@{}}\textbf{Paraphrase}: The failure of the Democrats to impeach Donald \\ Trump is a strong moment for our legislature which can get \\ back to its work helping the American people. Today we've been \\ able to tell the American people what we have known all along, \\ that Donald Trump was not guilty of these charges.\end{tabular} \\
\textbf{Wiki-Context Predicted}: Not Target | \textbf{DCF Prediction}: Target (correct)                                                                                                                                                                                                                                                                                                                                                                 & \textbf{Wiki-Context Predicted}: Yes | \textbf{DCF Prediction}: No (correct)                                                                                                                                                                                                                                                                                                            \\\hline & \\
\begin{tabular}[c]{@{}l@{}}\textbf{Tweet}: \#GeorgeFloyd \#BlackLivesMatter  \#justiceinpolicing \\ QT @OmarJimenez Former Minneapolis police officer Derek \\ Chauvin is in the process of being released from the Hennepin \\ County correctional facility  his attorney tells us. He is one of \\ the four officers charged in the death of George Floyd. He \\ faces murder and manslaughter charges.\end{tabular} & \begin{tabular}[c]{@{}l@{}}\textbf{Tweet}: Lots of honking and screaming from balconies. \\ Something must be going on.\end{tabular}                                                                                                                                                                                                                                \\
\textbf{Author}: Adriano Espaillat (Democrat)                                                                                                                                                                                                                                                                                                                                                                                         & \textbf{Affiliation}: Democrat                                                                                                                                                                                                                                                                                                                                      \\
\textbf{Event}: George Floyd protests                                                                                                                                                                                                                                                                                                                                                                                  & \textbf{Event}: Presidential election of 2020                                                                                                                                                                                                                                                                                                                       \\
\textbf{Entity}: Derek Chauvin                                                                                                                                                                                                                                                                                                                                                                                         & \begin{tabular}[c]{@{}l@{}}\textbf{Paraphrase}: I'm sure that the people are celebrating the \\ election results.\end{tabular}                                                                                                                                                                                                                                      \\
\textbf{Wiki-Context Predicted Sentiment}: Positive | \textbf{DCF Prediction}: Negative (correct)                                                                                                                                                                                                                                                                                                                                                            & \textbf{Wiki-Context Prediction}: No | \textbf{DCF Prediction}: Yes (correct)                                                                                                                                                                                                                                                                                                             \\ \hline
\end{tabular}

}
\caption{Examples where baseline model fails but DCF works}
\label{tab:tweet_ens_analysis}
\end{table*}

\section{Annotation Interfaces}
\begin{figure*}[tbh]
    \centering
    \includegraphics[width=\textwidth]{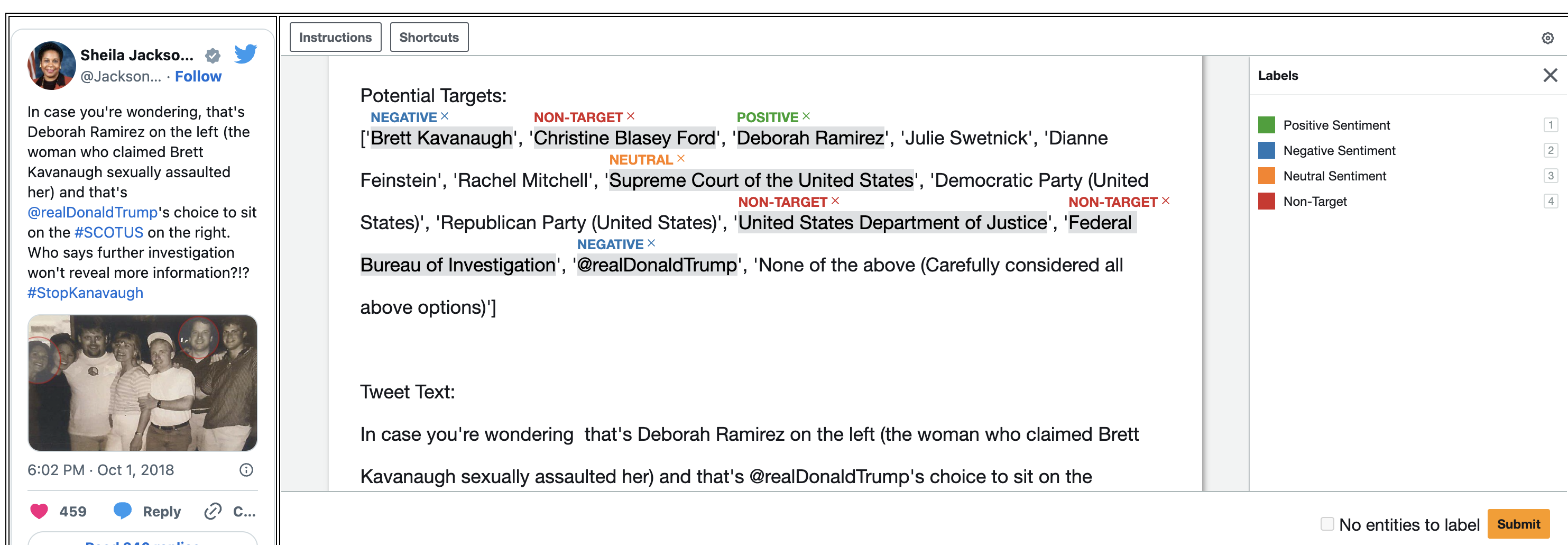}
    \caption{An example of \textit{Tweet Target Entity and Sentiment Annotation} GUI}
    \label{fig:enter-label1}
\end{figure*}

\begin{figure*}[tbh]
    \centering
    \includegraphics[width=\textwidth]{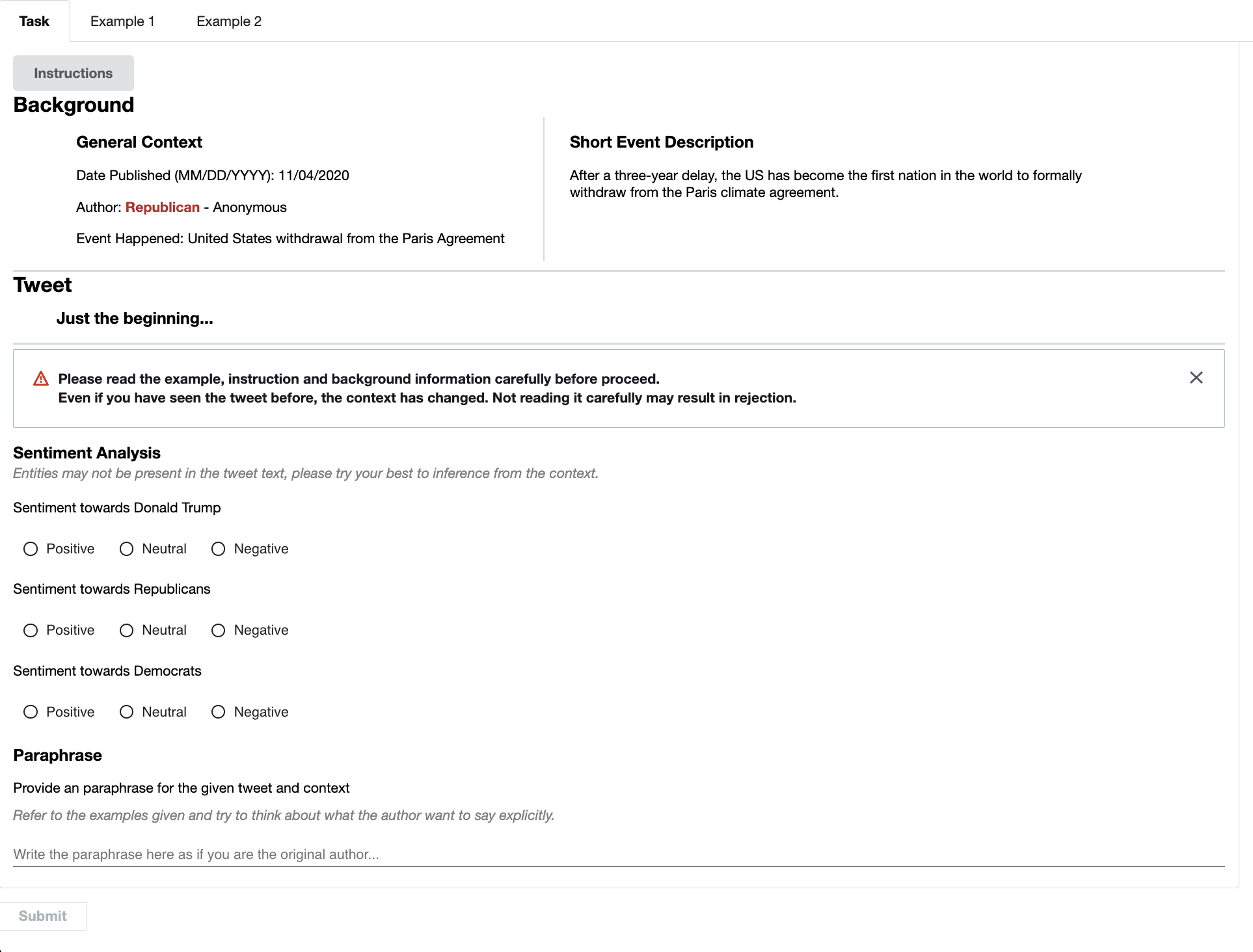}
    \caption{An example of \textit{Vague Text Disambiguation} GUI}
    \label{fig:enter-label2}
\end{figure*}

\end{document}